# POINT CLOUD GEOMETRY SCALABLE CODING WITH A QUALITY-CONDITIONED LATENTS PROBABILITY ESTIMATOR

*Daniele Mari*[1] *André F. R. Guarda*[4] *Nuno M. M. Rodrigues*[3,4] *Simone Milani*[1] *Fernando Pereira*[2,4]

[1] Università di Padova, Italy; [2] Instituto Superior Técnico – Universidade de Lisboa, Portugal; [3] ESTG – Politécnico de Leiria, Portugal; [4] Instituto de Telecomunicações, Portugal

**ABSTRACT**

The widespread usage of point clouds (PC) for immersive visual applications has resulted in the use of very heterogeneous receiving conditions and devices, notably in terms of network, hardware, and display capabilities. In this scenario, quality scalability, i.e., the ability to reconstruct a signal at different qualities by progressively decoding a single bitstream, is a major requirement that has yet to be conveniently addressed, notably in most learning-based PC coding solutions. This paper proposes a quality scalability scheme, named Scalable Quality Hyperprior (SQH), adaptable to learning-based static point cloud geometry codecs, which uses a Quality-conditioned Latents Probability Estimator (QuLPE) to decode a high-quality version of a PC learning-based representation, based on an available lower quality base layer. SQH is integrated in the future JPEG PC coding standard, allowing to create a layered bitstream that can be used to progressively decode the PC geometry with increasing quality and fidelity. Experimental results show that SQH offers the quality scalability feature with very limited or no compression performance penalty at all when compared with the corresponding non-scalable solution, thus preserving the significant compression gains over other state-of-the-art PC codecs.

## 1. INTRODUCTION

Point clouds (PCs) are a 3D visual representation format which is becoming increasingly popular in immersive, realistic, and interactive applications, notably virtual and augmented reality [1]. PCs are composed by a set of points, represented by their spatial coordinates (referenced as geometry), and, optionally, a set of associated features, notably color. To achieve a visually pleasant representation of a scene, PCs often contain millions of points, making the use of efficient coding algorithms a compelling requirement.

While multimedia coding based on conventional signal processing tools was dominant in the past, recent years have seen the emergence of learning-based coding approaches, with outstanding compression performance and compressed domain capabilities for many signal modalities. One good example is the recent specification of the JPEG AI learning-based image coding standard [2], which has achieved major compression performance gains regarding the best conventional image coding solutions, notably the Versatile Video Coding (VVC) [3] Intra mode codec, as well as compressed domain processing benefits for tasks such as classification, segmentation and recognition [4].

In the multimedia signal coding arena, PC coding is a particularly challenging task, due to the unstructured nature of the PC data, and the commonly heterogeneous point density, in contrast to the regularly structured images. To address the interoperability requirements of real applications, the MPEG standardization group has pioneered the development of standards for PC coding. The geometry-based Point Cloud Compression (G-PCC) and Video-based Point Cloud Compression (V-PCC) [5] MPEG standards, respectively for static and dynamic PC coding, are based on conventional signal processing coding techniques. More recently, the JPEG standardization group started developing a learning-based PC coding standard, called JPEG Pleno Point Cloud Coding (PCC) [6], targeting static PC coding. These efforts envision the development of a learning-based PC coding standard, offering competitive compression performance for interactive human visualization and effective compressed domain performance for machine-related computer vision tasks [7]. The so-called JPEG Pleno PCC Verification Model (VM) [8], from this point onwards referred to as JPEG PCC, is the current version of the future JPEG PCC standard, and has an associated software implementation. JPEG PCC is one of the best performing codecs for static PC geometry, even when compared with G-PCC and V-PCC in Intra mode [9], especially for dense PCs.

The JPEG PCC geometry coding model architecture consists of two main modules. The first is an autoencoder (AE), which is used to transform the input PC geometry into a more compact representation, commonly referred as the latent representation. The second module is a variational autoencoder (VAE), which is used to estimate a prior to more efficiently entropy code the latent representation, thus exploiting a so-called hyperprior [10]. The JPEG PCC encoder and decoder models are jointly trained using an end-to-end approach and a loss function that enforces a given target RD trade-off, using a specific rate control parameter. Therefore, to achieve different target rates/qualities, it is required to train and store different and independent JPEG PCC encoder/decoder models, which produce different and independent bitstreams.

In a real-world transmission scenario, heterogenous receivers operate under diverse conditions related to network, hardware or display-related limitations: as a consequence, they need to consume bitstreams at different rate/quality levels. The simplest way to fulfil these requirements would be transmitting multiple independent coded streams, one for each quality level, using a so-called *non-scalable coding* paradigm. Alternatively, a *scalable coding* paradigm may be adopted by creating a single, appropriately structured stream that can simultaneously serve all (heterogeneous) receivers. The key feature of quality scalable codecs is to encode and decode each layer exploiting some information from the previous layers (already available at both the encoder and decoder), thus successively enhancing the signal quality, starting from the so-called *base layer*. The exploitation of the previously available information, which may happen in multiple ways, notably in the spatial/temporal or compressed domains, generally comes at the cost of an additional bitrate overhead when compared to equivalent non-scalable solutions. As for other media modalities, the growing heterogeneity of the consumption conditions has recently increased the need for scalability in PC codecs, notably for automatic machine vision [11] and VR/gaming applications [12]. As a response to these increasing needs, current PC standards, including JPEG PCC, have included in their requirements various types of scalabilities, notably geometry quality scalability [13].

In this context, this paper proposes a novel quality scalable PC geometry coding scheme, called Scalable Quality Hyperprior (SQH), designed on top of the JPEG PCC codec. This novel method exploits the correlation between the JPEG PCC latents, independently obtained with different coding models for different target rates/qualities, to create a layered, scalable bitstream where each reconstructed signal builds on top of a lower rate/quality layer. This is done by using a Quality-conditioned Latents Probability Estimator (QuLPE), which exploits the latents from a previous layer to estimate a better probability distribution to more efficiently encode the latents of a higher quality layer.

The advantage of the QuLPE estimator model is that it can handle latents at all possible rate/quality levels, meaning that scalability is achieved by adding a single neural network to JPEG PCC and not different models for different rates/qualities. Unlike other scalable coding solutions, experimental results show that SQH provides extra flexibility with very limited or even no RD compression penalty with respect to its non-scalable counterpart. Indeed, for some specific configurations, SQH is even able to achieve marginal RD performance gains regarding the non-scalable benchmark, which is a remarkable achievement compared to available scalable coding solutions for other modalities.

Additionally, since SQH acts directly on the latents' domain, it offers other advantages when compared with scalable coding solutions working in the spatial domain. Firstly, spatial domain approaches generally rely on computing residuals which, for PCs, tend to be very sparse and hard to code: this problem does not apply in the latent domain. Secondly, only the PC at the target quality needs to be fully decoded, without requiring to fully decode the previous layers, thus considerably lowering the decoding complexity/time. Finally, and more importantly, even if in this paper the SQH model was implemented and demonstrated on top of JPEG PCC, this framework may be easily generalizable to any autoencoder-based codec and other types of signals, e.g., images.

The rest of the paper is organized as follows: Section 2 reviews the current state-of-the-art in PC coding, with particular focus on scalable solutions. Section 3 reviews the JPEG PCC codec, which serves as the reference and base layer codec for this paper. Section 4 proposes the novel SQH quality scalable coding procedure and the underlying QuLPE model architecture. Finally, Section 5 reports and analyses the most relevant experimental results and Section 6 discusses future work directions.

## 2. RELEVANT BACKGROUND WORK

This section provides a brief review of the state-of-the-art in PC coding with particular focus on scalable solutions. PC coding solutions can be simply divided into conventional signal processing-based solutions [5] and emerging DL-based solutions [8][14][15].

The G-PCC and V-PCC MPEG standards [5] are the most important references for the best performing signal processing-based PCC solutions. G-PCC exploits an octree representation to efficiently code the PC geometry while V-PCC projects the PC signal to the image domain and uses well established image/video codecs such as HEVC or VVC [3] to compress them.

Recently, a new type of coding solutions based on DL models have stormed the PCC arena. The DL-PCSC [16] codec partitions the latents in groups in a way that allows to encode and decode the PC at increasing qualities by progressively transmitting more latent values. Whenever a partition is not transmitted, the decoder will use zeroes for these latents instead. However, this strongly limits the flexibility of the network since it constrains the latent space around the zero value, reducing its flexibility and introducing a very high scalability coding penalty. Additionally, previous works [10] have shown that it is convenient to have a larger latent space to give more freedom to the network to learn how to allocate the rate; thus, an approach like DL-PCSC, tends to be more limiting in terms of RD performance.

Another solution supporting quality scalability is Grasp-Net [17]. Grasp-Net proposes a DL-based enhancement layer building on top of G-PCC which acts as the base layer. Since scalability is not the main aim of Grasp-Net, the codec has limited flexibility since it only supports a maximum of two qualities per bitstream. Furthermore, the base layer is usually encoded at a very low precision to allow for a more efficient use of the enhancement layer.

One other relevant example is SparsePCGC [15] which encodes a sparse voxelized PC by down-sampling it multiple times while building bitstreams that allow to losslessly reverse the down-sampling operation. The receiver is thus able to up-sample the PC and either request the information required to losslessly up-sample the PC at the considered layer or use a lossy thresholding strategy to decide which voxels are occupied and which ones are not. Sparse-PCGC has a very good coding performance compared with other quality scalability PC codecs. Nevertheless, the scalability approach is highly dependent on the Sparse-PCGC encoding strategy and generally cannot be applied to other codecs.

## 3. JPEG PCC VERIFICATION MODEL

This section presents the codec adopted as the basis for the proposed SQH scheme, which is the latest version of the JPEG Pleno PCC VM [8]; this codec uses a learning-based coding approach to code both the geometry and color components [18]. For geometry coding, JPEG PCC uses a DL model consisting of an autoencoder with a mean and scale hyperprior [19]. Additionally, two more tools can be used for geometry coding, which have clear compression performance advantages for sparse PCs and PC coding at lower rates: a down-sampling module using a sampling factor (SF) parameter; and a DL-based super-resolution (SR) module to enhance the reconstructed quality when down-sampling is performed. JPEG PCC encodes the PC color information by projecting texture patches onto an image (similarly to V-PCC [9]) which is then coded using the emerging JPEG AI codec [2].

The latest version of the JPEG PCC VM (version 3.0) geometry codec adopts a sparse tensor representation [20], which leads to improvements both in terms of computational complexity and RD performance. In such a representation, PCs are represented as tuple $x = (x_C, x_F)$ where $x_C$ are the points' coordinates and $x_F$ the corresponding features, which are initially set to the value "1" to indicate that the voxel is not empty.

Since the focus of this paper lies in geometry coding, the remaining of the text will focus on this component only. The overall DL-based JPEG PCC geometry coding model architecture is shown in Fig.1. The coding/decoding procedures are now briefly reviewed.

At the sender, the processing operations proceed as follows:

S1. Compute the latents $y$, composed by their coordinates $y_C$ and feature values $y_F$, from the input PC $x$, using the analysis transform ($\mathcal{G}_a$) as $y = \mathcal{G}_a(x)$.

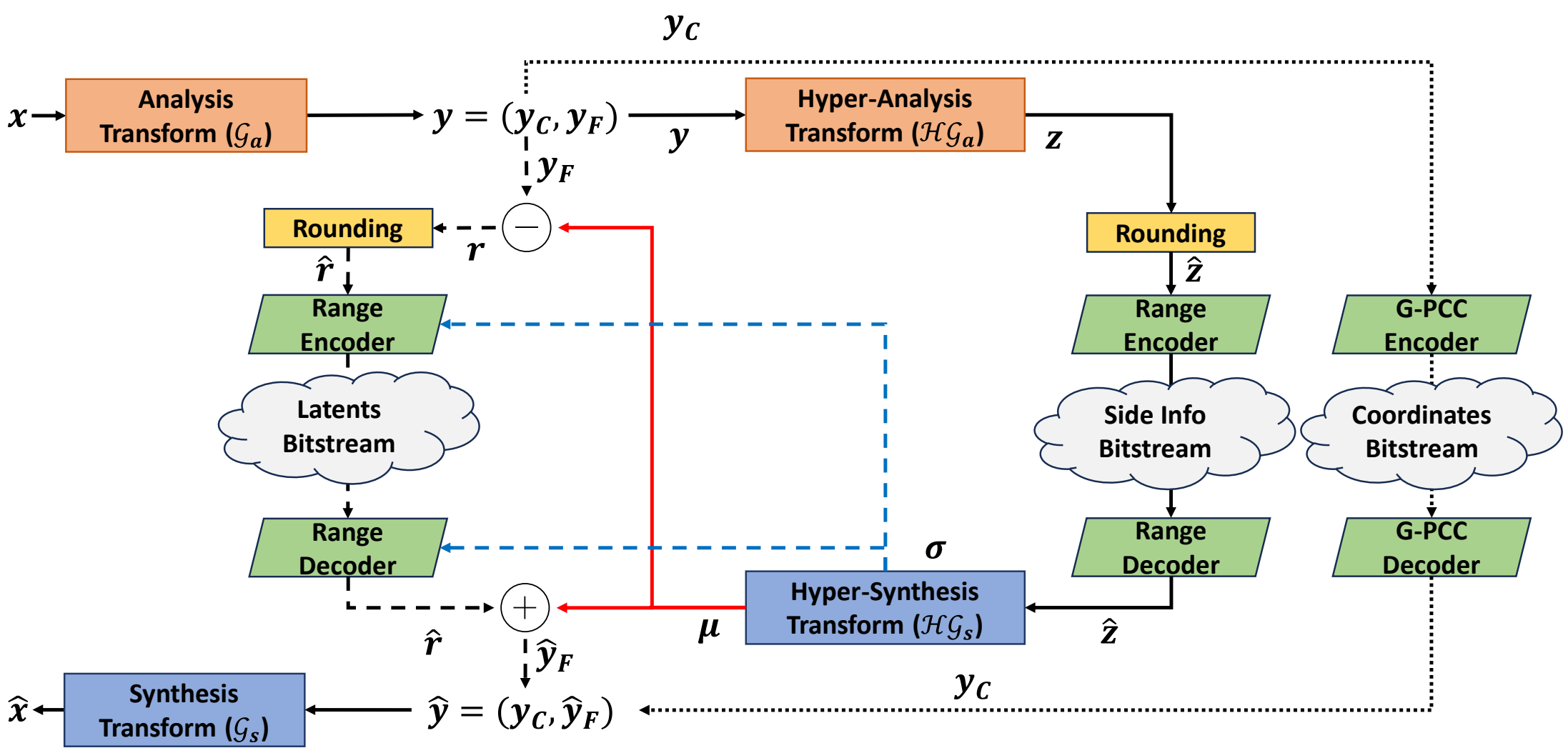

Fig. 1. JPEG Pleno PCC VM geometry coding architecture.

S2. Encode the latents' coordinates $y_C$, using G-PCC octree, into the coordinates bitstream.

S3. Compute the hyper-latents, $z$, from the latents, using the hyper-analysis transform model ($\mathcal{HG}_a$), as $z = \mathcal{HG}_a(y)$. Then transform $z$ into $\hat{z}$, by rounding to the closest integer.

S4. Entropy code the hyper-latents using a range encoder according to a fully factorized prior $p(\hat{z})$, and transmit them as a side information bitstream.

S5. Feed the decoded hyper-latents, $\hat{z}$, to the hyper-synthesis transform ($\mathcal{HG}_s$) to estimate the parameters of the Gaussian modeling the conditional probability distribution $p(y|\hat{z})$, notably the mean and standard deviation as $\mu, \sigma = \mathcal{HG}_s(\hat{z})$.

S6. Entropy code the latents' features, $y_F$, using the estimated values for $\mu$ and $\sigma$, thus finally producing the latents bitstream.

The receiver reconstructs the PC as follows:

R1. Losslessly decode the latents' coordinates $y_C$ and hyper-latents $\hat{z}$ from the coordinates and side information bitstreams, respectively.

R2. Compute the mean, $\mu$, and standard deviations, $\sigma$, using the hyper-synthesis transform, as $\mathcal{HG}_s(\hat{z})$.

R3. Entropy decode $\hat{y}_F$ from the latents' bitstream using $\mu$ and $\sigma$.

R4. Feed the decoded latents, $\hat{y} = (\hat{y}_F, y_C)$, to the synthesis transform ($\mathcal{G}_s$), to reconstruct the PC as $\hat{x} = \mathcal{G}_s(\hat{y})$.

The encoder and decoder models are trained with an end-to-end approach, using the previously explained operations, with one exception: quantization is substituted with a differentiable approximation to make the model fully differentiable. The coding model is trained end-to-end using a standard RD loss function as:

$$\mathcal{L}(x, \hat{x}, y, z) = \mathcal{D}(x, \hat{x}) + \lambda \cdot \mathcal{H}(y, z), \tag{1}$$

where $\mathcal{D}(\cdot,\cdot)$ is the distortion, measured as the focal loss [21], $\mathcal{H}(\cdot,\cdot)$ is the entropy under the estimated probability distributions $p(\hat{z})$ and $p(y|\hat{z})$, and $\lambda$ is a trade-off parameter to control the target compression ratio.

Generally, one model is trained for each RD point corresponding to one value of $\lambda$. In JPEG PCC, five different coding models are trained to support the defined range of rates/qualities. The training procedure is carried out in a curricular manner by sequentially spanning the chosen values of $\lambda \in \{0.05, 0.025, 0.01, 0.005, 0.0025\}$, progressively moving from the lowest value (highest rate/quality) to the highest one (lowest rate/quality). In the following sections, these rate points and their related trained models will be denoted by the index $i \in \{1, \dots, 5\}$, with $i = 1$ corresponding to $\lambda = 0.05$ (lowest rate/quality) and $i = 5$ to $\lambda = 0.0025$ (highest rate/quality).

## 4. SCALABLE QUALITY HYPERPRIOR

This section proposes the novel Scalable Quality Hyperprior scheme which allows the JPEG PCC codec to support quality scalability. Offering scalability in the spatial domain implies decoding the geometry for each rate/quality level, thus increasing the final decoding complexity and the sparsity of the residuals, so working in the latents domain is advantageous. However, the latent spaces produced by autoencoder models trained independently with different $\lambda$ values tend to be very different from one another, meaning that these compressed representations do not traditionally favor the construction of a layered bitstream as required for scalability. However, preliminary observations have shown that the latents computed using the five JPEG PCC coding models are clearly correlated. The next subsection reports these observations and the general intuition behind the proposed SQH scalability approach. This is followed by a detailed description of the SQH coding scheme and QuLPE model.

### 4.1. Preliminary observations and intuition

Experimental evaluations have shown that the latent spaces generated by the multiple JPEG PCC coding models are relatively aligned. This can be concluded from in Fig. 2, which shows that the average cosine similarity between the latents produced by the different coding models for the same input sample $x$ is, on average, very high. As previously mentioned, this was not expected and may likely result from the adoption of sequential training in JPEG PCC. This learning strategy, which consists in training the i-th model (producing latents $y_{\mathrm{i}}$) starting from the weights of model i+1 (producing latents $y_{\mathrm{i+1}}$), i.e. models for lower rates from models for higher rates, was originally proposed to reduce the training time. However, these observations suggest that, since the latents show a strong correlation across different rate/qualities, it is possible to estimate a good entropy model for the higher quality representations from the lower quality ones, making entropy coding more efficient.

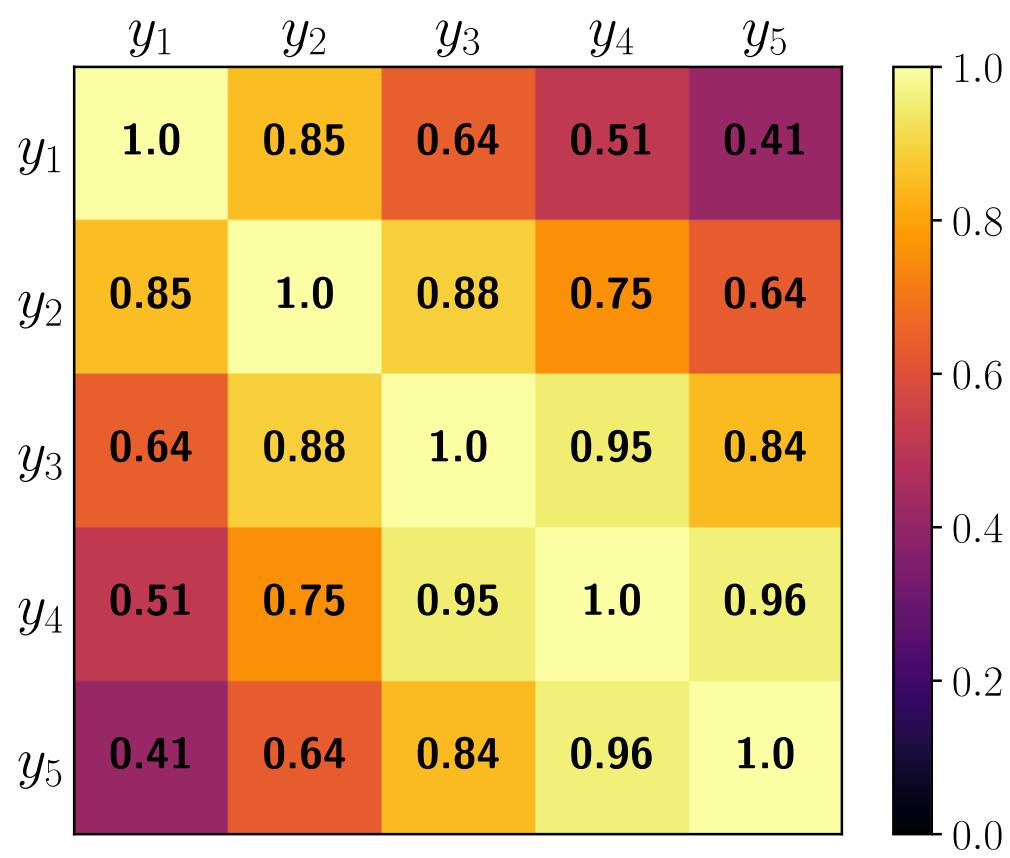

Fig. 2. Average cosine similarity between corresponding latents produced by the five different trained JPEG PCC coding models.

### 4.2. SQH coding scheme

Based on the previous intuition, this section proposes the SQH coding scheme for quality scalable PC geometry coding, whose diagram is represented in Fig. 3. In a nutshell, the base layer creates the three bitstream components produced by JPEG PCC, i.e., latents, side information and coordinates bitstreams, while the following enhancement layers (in Fig. 3 only one enhancement layer is represented for simplicity) use the SQH bitstream, i.e., the information allowing to decode $\hat{y}$ at progressively higher qualities. The SQH bitstream corresponds to the JPEG PCC latents bitstream when using SQH for the enhancement layers. Additionally, no coordinates bitstream is needed after the base layer since it does not change for subsequent layers (assuming a constant sampling factor, SF) and no side bitstream with hyper-latents is required since the latents decoded by the base layer act as the new side information.

More specifically, assuming that a base layer version of a PC, $x$, is available at both the sender and receiver at a given quality, $i$ (by using the JPEG PCC coding procedure explained in Sec. 3), the sender performs the following operations to obtain the next (higher rate/quality) layer bitstreams:

S1. Use the JPEG PCC coding model at a higher target quality $i+n$, (with $i>0, n>0$ and $i+n \leq 5$), to compute the new latents $y_{i+n}$.

S2. Estimate the means and standard deviation values for the latents above using the DL-based QuLPE model (described in the next section) as $\mu_{i+n}, \sigma_{i+n} = QuLPE(\hat{y}_i, i, i+n)$, assuming that the latents are distributed as independent Gaussians, $P(y_{i+n}|\hat{y}_i)$.

S3. Entropy encode $y_{i+n}$ using $\mu_{i+n}, \sigma_{i+n}$ , thus obtaining the SQH bitstream.

The entropy modelling procedure is similar to the one carried out in JPEG PCC since, for both cases, some hyperprior is exploited to estimate a Gaussian prior for the latents. The only difference is that in the SQH solution the hyperprior is $\hat{y}_i$, i.e., the previously decoded latents, while in JPEG PCC the $i+n$ model hyper-latents $\hat{z}_{i+n}$ are used instead (now not needed).

On the other hand, the receiver, which has access to the base layer information $\hat{y}_i$, recovers the higher rate/quality PC using the following decoding steps:

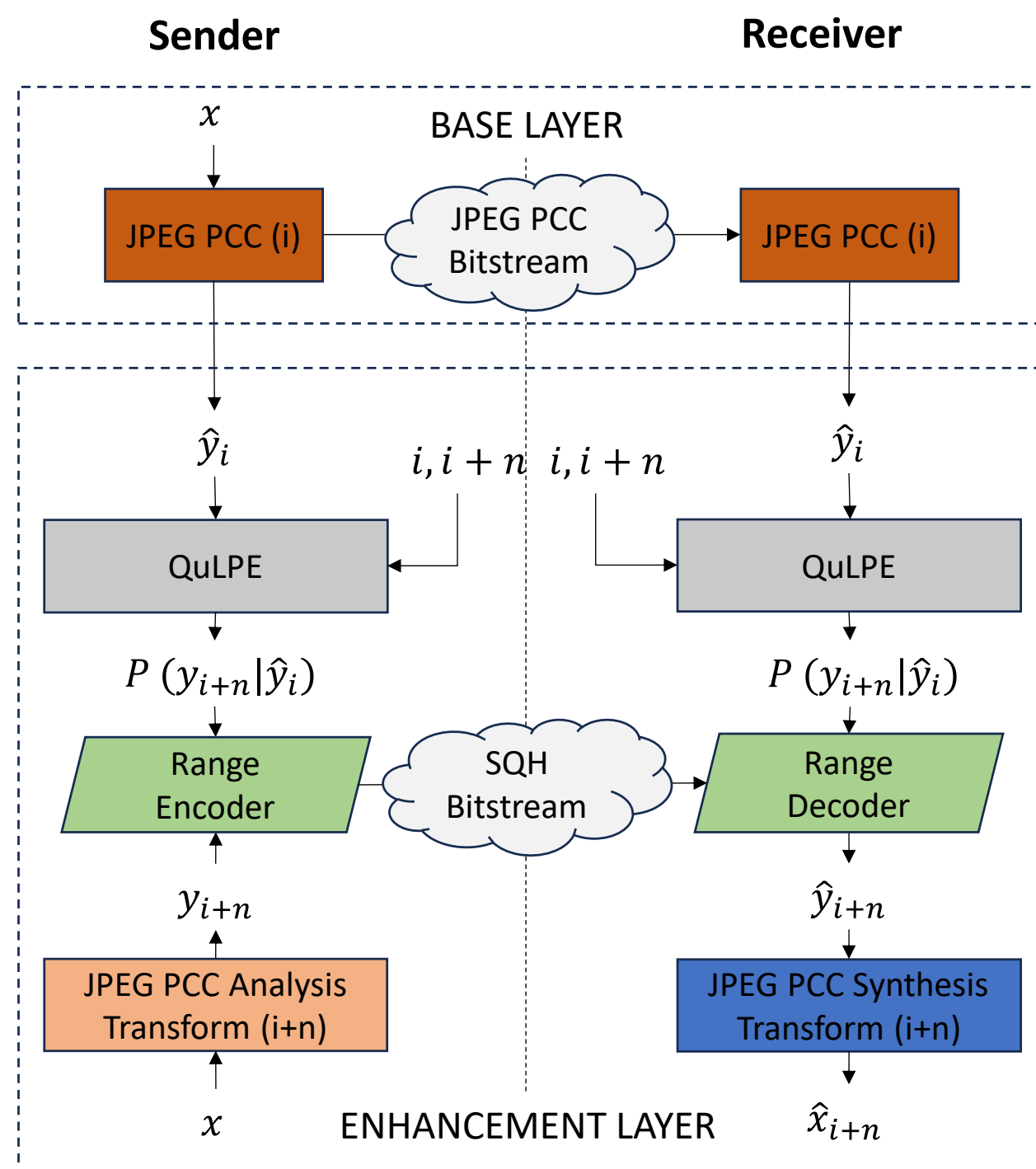

Fig. 3. Scalable Quality Hyperprior coding scheme.

R1. Estimate $\mu_{i+n}, \sigma_{i+n} = QuLPE(\hat{y}_i, i, i+n)$ from the base layer information, $\hat{y}_i$, using the proposed QuLPE model.

R2. Decode the $\hat{y}_{i+n}$ enhancement layer latents using a range decoder applied to the SQH bitstream.

R3. Reconstruct the final PC using the JPEG PCC decoder applied to the enhancement layer latents, thus obtaining $\hat{x}_{i+n} = \mathcal{G}_{s,i+n}(\hat{y}_{i+n})$.

Naturally, if more than two scalability layers are required, it is possible to repeat the SQH encoding and decoding steps as many times as needed, with the key difference that the latents newly decoded after running the enhancement layer should be used as the new base layer for the next layer.

### 4.3. QuLPE model architecture and training

This subsection proposes the novel QuLPE model, which is used to estimate the mean and standard deviation parameters, $\mu_{i+n}, \sigma_{i+n}$ which are the parameters for the Gaussian prior $P(y_{i+n}|\hat{y}_i)$.

Since JPEG PCC is trained with five different values of $\lambda$, the QuLPE model must estimate $P(y_{i+n}|\hat{y}_i)$ for all possible values of $i$ and $n$ to achieve full scalability. Given the restriction $1 \leq i < i+n \leq 5$, the number of possible combinations is very high and increases quadratically with the number of trained models. For the particular case of JPEG PCC, this would require training ten SQH models, which would have to be available at the receiver side to enable decoding all possible layered streams. To avoid this memory burden, a single QuLPE model is designed to handle all possible combinations of rates/qualities with a single NN. However, to perform a meaningful estimation of the required probability distribution, the model needs to know the rates/qualities of the source ($\hat{y}_i$) and the target ($y_{i+n}$) latents. For this reason, QuLPE is conditioned on the source and target qualities by feeding it with the quality indices $i, i+n$, in addition to $\hat{y}_i$.

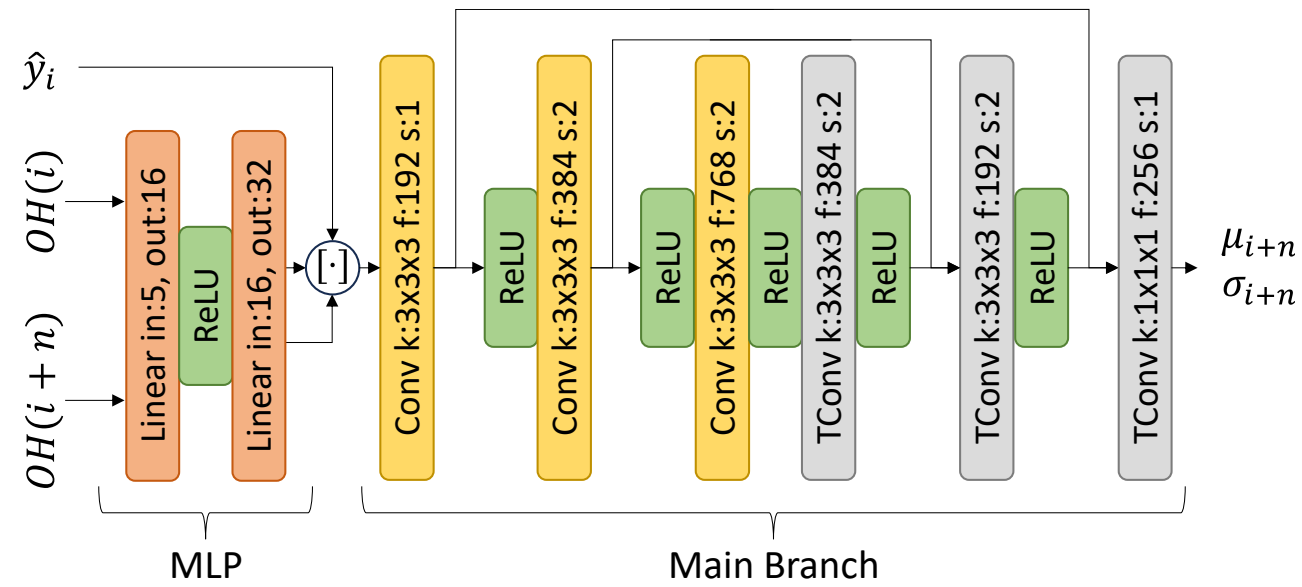

Fig. 4. QuLPE model architecture.

The proposed QuLPE model architecture is represented in Fig. 4. As it is possible to see, the two indexes $i, i+n$ are first transformed into some more expressive representations, called embeddings that are computed by processing the one-hot (OH) representation of the indexes with a Multi-Layer Perceptron (MLP). These embeddings are then concatenated with the source latents, $\hat{y}_i$, and fed to the main network branch that will finally predict the probability model parameters, $\mu_{i+n}, \sigma_{i+n}$.

More specifically the input to the main QuLPE branch is computed as

$$y_{in} = [\hat{y}_i, f_\theta(OH(i)), f_\theta(OH(i+n))], \quad (2)$$

where $f_\theta(\cdot)$ is the aforementioned MLP, $OH(\cdot)$ is the one hot encoding function, and $[\cdot]$ denotes concatenation.

At this point, $y_{in}$ is used as input to the main network branch which resembles a U-Net [22] since it has an hourglass-like shape with skip connections between corresponding layers with the same features size. This favors better entropy model prediction since no information is lost at the bottleneck in the middle. While skip connections cannot be used in the most common hyperprior models since all information shared from the encoder to the decoder needs to be transmitted, this is not the case for the novel QuLPE since both the sender and receiver require only the base layer latents to compute the output. The SQH model is implemented using sparse convolutions, since they are better suited for PC processing [1] and better fit the JPEG PCC software.

The training and validation datasets follow the JPEG Pleno PCC Common Training and Testing Conditions (CTTC) [23], with the only difference being that block size is set to 128 instead of 64. The QuLPE model is trained using a dataset containing the latent representations produced by JPEG PCC for the CTTC PC training dataset. Each $128 \times 128 \times 128$ PC block is encoded using the five JPEG PCC models (with different rate/quality levels) with SF equal to 1, thus obtaining the five latent representations for each block. During training, at each gradient update step, a tuple $(i, i+n)$ is selected for each training PC block with uniform probability, and the corresponding latents $\hat{y}_i, y_{i+n}$ are loaded accordingly. During validation, all possible combinations of qualities are evaluated to guarantee consistency between the validation loss across all epochs.

The SQH model is trained to minimize the loss function

$$\mathcal{L}(y_{i+n}, \hat{y}_i) = \mathcal{H}(y_{i+n}|\hat{y}_i) \quad (3)$$

where $\mathcal{H}(y_{i+n}|\hat{y}_i)$ is the entropy of the latents under the estimated prior $P(y_{i+n}|\hat{y}_i)$. No distortion component is used since the aim of the model is to transmit $y_{i+n}$ with as few bits as possible. The model is trained using the Adam optimizer using a learning rate equal to $10^{-3}$, exponentially decaying by a factor of 10 after 7 epochs without improvement. Additionally, early stopping terminates the training when the validation error does not decrease for more than 10 epochs.

## 5. PERFORMANCE ASSESSMENT

This section first presents the experimental setup, followed by the SQH performance in comparison with appropriate benchmarks.

### 5.1. Experimental setup

The experimental evaluation was conducted according to the JPEG Pleno PCC Common Training and Testing Conditions (CTTC) [23]. The test dataset includes 25 PCs with varying content and characteristics. The quality metric used for the evaluation of the RD performance is PSNR D1. Results for PSNR D2 were omitted since their behavior is similar to PSNR D1. The rate is measured in terms of bits-per-(original)point (bpp).

The RD performance for the JPEG PCC quality scalable codec using SQH (from now on referred to as SQH) was compared with five relevant benchmarks, three non-scalable codecs and two scalable ones, notably:

1. G-PCC in Octree mode [5];
2. V-PCC in Intra mode [9];
3. JPEG-PCC, with SF=1 and no SR, referred to as JPEG PCC;
4. JPEG PCC independent, an ad-hoc algorithm for JPEG PCC-based scalability;
5. Grasp-Net [17].

Results for the Sparse-PCGC codec have not been included since the corresponding source code and trained models are not publicly available.

In the so-called JPEG PCC independent benchmark, all layers are encoded separately, i.e., without reusing previously transmitted information, which significantly raises the rate. In JPEG PCC, using SF=1 and no SR as coding parameters is not optimal for all PCs; in particular, for sparser PCs, tuning these two parameters might yield better RD performance. However, finding the optimal parameters for each PC would require an extensive search which is very time consuming. For this reason, SF=1, and thus no SR, were chosen for simplicity as the parameters used to compress all PCs, which is generally a reasonable choice. Additionally, this helps to evaluate the difference between JPEG PCC and SQH using the same coding conditions since again SF=1 was used for JPEG SQH. This is also required since SQH, in its current version, does not support the subsampling tool unless the same SF is used for all the target rates/qualities. Supporting the usage of different SFs for different target rates/qualities would require a new strategy for latent up-sampling, which is outside the scope of this paper and will be considered in future work.

For the MPEG standards G-PCC and V-PCC, the reference software version 20 was used. For Grasp-Net, the source code and trained models from the original repository [17] were used. For JPEG PCC, the Verification Model version 3.0 was used.

### 5.2. RD performance for JPEG PCC-based solutions

This subsection compares the RD performance for the proposed quality scalable SQH solution with the performance for all the other JPEG PCC-based coding benchmarks.

The SQH compression performance was assessed for different scalability configurations defined by the rate/quality level of the base layer and the enhancement layers. The notation SQH($i_1$,$i_2$,...,$i_k$) is used to refer to a configuration which adopts as base layer the quality level $i_1$, and as enhancement layers the following $k-1$ layers, corresponding to quality levels $i_2$,...,$i_k$. This configuration results in a quality scalable bitstream $b = [b_{i_1}, b_{i_2}, \dots, b_{i_k}]$ which may be used to sequentially decode $\hat{x}_{i_1}, \dots, \hat{x}_{i_k}$, thus defining a RD curve with $k$ RD points. The distortion for the $t^{th}$ RD point, with $1 \leq t \leq k$, is computed as the distortion of PC $\hat{x}_{i_t}$, while the rate is $\mathcal{R}(t) = \mathcal{R}(b_{i_1}) + \ \dots + \mathcal{R}(b_{i_t})$.

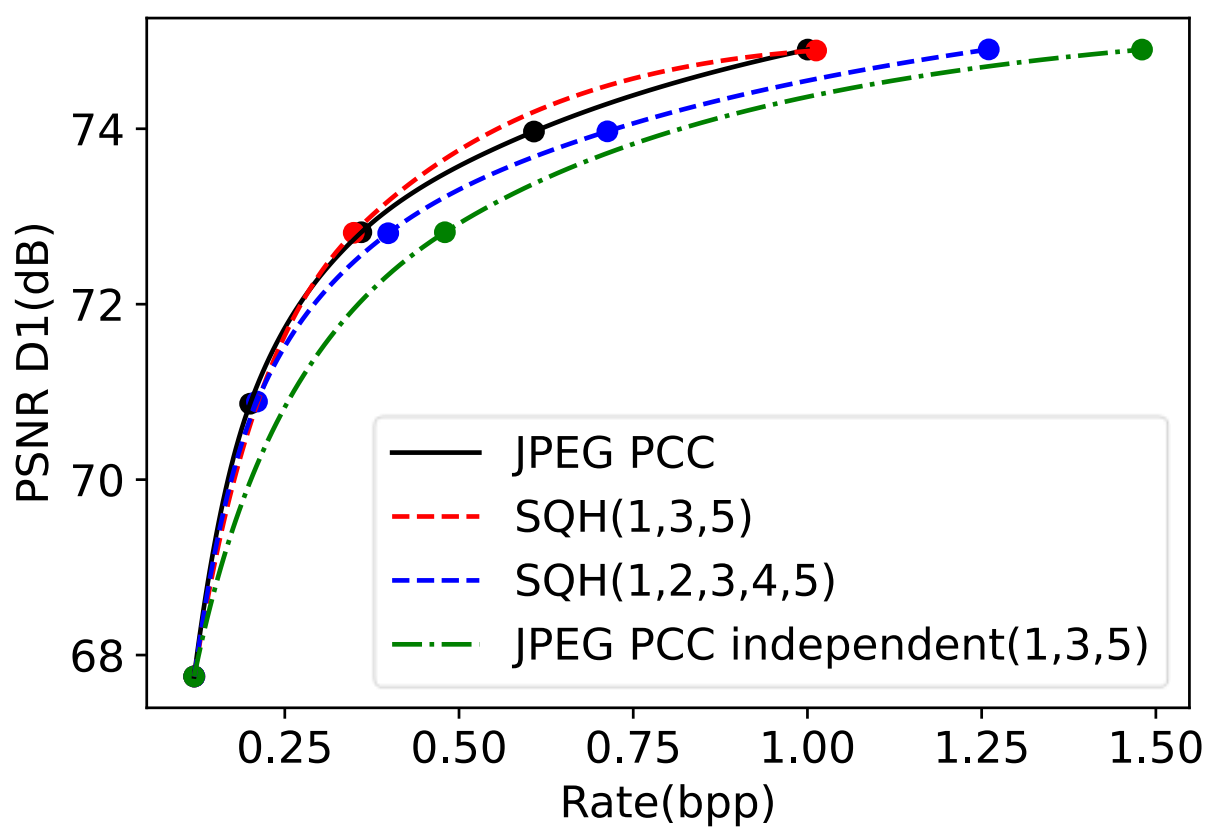

Fig. 5. PSNR D1 RD performance comparison for JPEG PCC-based solutions (average over the test dataset).

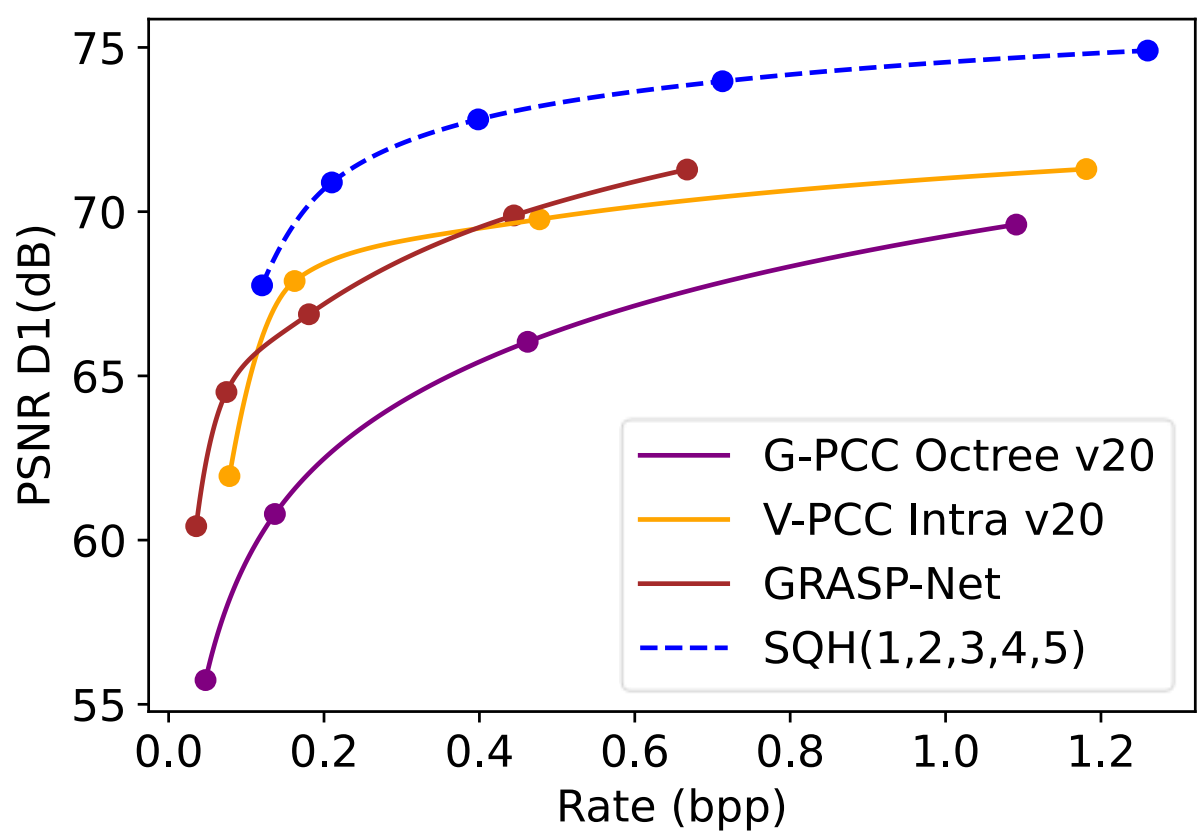

Fig. 6. PSNR D1 RD performance comparison with other benchmarks (average over the test dataset).

The experimental RD performance is shown in Fig. 5 and include two SQH-based solutions, notably SQH(1,2,3,4,5) and SQH(1,3,5). SQH(1,2,3,4,5) corresponds to the case where full scalability is offered, i.e., involving all the five target rates/qualities, while SQH(1,3,5) covers a more limited scalability setting where only the lowest, middle, and highest rates/qualities are decodable.

The RD performance shown in Fig. 5 leads to the following three main observations:

a) When comparing the SQH solutions with JPEG PCC, it is possible to notice that scalability comes with a very limited or almost no cost in terms of RD performance.
b) As expected, SQH(1,2,3,4,5) leads to slightly higher rates than SQH(1,3,5) due to the fact that the former allows decoding the PC at two more qualities (2 and 4) than the latter; so the additional flexibility is paid with some additional rate.
c) When comparing SQH(1,3,5) with the corresponding JPEG PCC independent (1,3,5) solution, it is possible to observe significant performance gains. These directly measure the positive impact of exploiting the information in previous layers, instead of transmitting all the layers independently.

In conclusion, the RD performance cost that has to be paid for the extra flexibility offered by the SQH scalability is quite limited, making it an appealing solution since this important feature comes almost for free, contrary to past solutions for other modalities.

On average, it is even possible to notice a small gain for SQH(1,3,5) over JPEG PCC, notably for $i = 3$, suggesting that, in some cases, the base layer latents, $\hat{y}_i$, might be better side information for the target layer than the hyper-latents $\hat{z}_{i+n}$. As a matter of fact, at rates/qualities 3, 4, 5, encoded with configurations SQH(1,3), SQH(1,4), SQH(1,5), SQH saves on average 3.0%, 5.2%, 4.8% bpp for qualities 3, 4, 5 respectively w.r.t JPEG PCC. This suggest that, at higher bitrates, the standard hyperprior might not be the best possible side information; naturally, these gains are not guaranteed to hold if more complex entropy models, e.g., autoregressive models, are adopted in JPEG PCC.

### 5.3. RD performance benchmarking

This subsection compares the RD performance for the proposed quality scalable SQH solution, notably SQH(1,2,3,4,5), with the non-JPEG PCC-based benchmarks, notably the non-scalable G-PCC and V-PCC solutions and the scalable GRASP-Net solution. The RD performance displayed in Fig. 6 shows that SQH(1,2,3,4,5) provides full scalability over the target rates/qualities while offering significant RD performance gains compared with conventional non-scalable codecs, in this case G-PCC and V-PCC, as well as with the partially scalable learning-based GRASP-net codec.

Overall, the RD performance results show that the SQH scalability has an acceptable RD performance loss when compared with JPEG PCC, while maintaining a clear advantage over other state-of-the-art solutions, which have no or limited scalability. Furthermore, the RD performance loss depends on the required scalability flexibility. Nevertheless, it can be seen that if less flexibility is required, notably with the SQH(1,3,5) configuration,, then it is possible to match the RD performance of JPEG PCC.

## 6. CONCLUSIONS

This paper proposes a quality scalability scheme for JPEG-PCC, labelled as SQH, which offers a scalable bitstream with little to no RD performance loss with respect to the corresponding non-scalable solutions. Additionally, since SQH scalability is implemented in the latent space, it allows avoiding some of the usual downsides of spatial domain scalability, such as the sparsity of the residuals and the need to decode the PC at every target quality. Finally, contrary to SparsePCGC, since scalability is provided thanks to an extra component, i.e. the QuLPE model, JPEG PCC can still be used in a non-scalable way if this feature is not required.

Future work will focus on integrating the SQH scheme in JPEG PCC while allowing different sampling factors. Moreover, the effectiveness of the SQH scheme in the context of other autoencoder-based codecs such as JPEG AI and other learning-based image codecs will be assessed. Finally, preliminary results have shown that latents from the previous layers act, in some cases, as better side information than the hyper-latents, thus suggesting a possible path towards the improvement of current hyper-priors.

## 7. ACKNOWLEDGEMENTS

This work was partially supported by the European Union under the Italian National Recovery and Resilience Plan (NRRP) of NextGenerationEU, partnership on “Telecommunications of the Future” (PE00000001—program “RESTART”). Daniele Mari’s activities were supported by Fondazione CaRiPaRo under the grants “Dottorati di Ricerca” 2021/2022. This research was funded in part by the Fundação para a Ciência e a Tecnologia, I.P. (FCT, Funder ID = 50110000187) under Grant PTDC/EEI-COM/1125/2021.